\documentclass[11pt]{article}
\usepackage[margin=1in]{geometry}

\usepackage[utf8]{inputenc}
\usepackage{microtype}           
\usepackage{graphicx}            
\usepackage{amsmath, amssymb}   
\usepackage{booktabs}            

\usepackage{abstract}
\usepackage{xcolor}
\usepackage{tabularx}
\usepackage{enumitem}
\usepackage[numbers,super,compress]{natbib}
\usepackage{url}

\usepackage[
  colorlinks=true,   
  linkcolor=blue,     
  citecolor=blue,     
  urlcolor=blue       
]{hyperref}

\newcommand{\stratuntrained}{\textsc{untrained}}
\newcommand{\stratuniform}{\textsc{random}}
\newcommand{\stratcdr}{\textsc{cdr}}
\newcommand{\stratspan}{\textsc{span}}
\newcommand{\stratstruct}{\textsc{structure}}
\newcommand{\stratstructLR}{\textsc{structure-lr}}
\newcommand{\stratint}{\textsc{interface}}
\newcommand{\stratgerm}{\textsc{germline}}
\newcommand{\stratinter}{\textsc{intersection}}
\newcommand{\strathyb}{\textsc{hybrid}}
\newcommand{\strathybst}{\textsc{hybrid-stretched}}
\newcommand{\strathybrv}{\textsc{hybrid-reverse}}
\newcommand{\strathybws}{\textsc{hybrid-warmstart}}
\newcommand{\strathybw}{\textsc{hybrid-weighted}}
\newcommand{\strathybpb}{\textsc{hybrid-perbatch}}

\newcommand{\strathybrand}{\textsc{hybrid-random}}

\title{%
  Learning Task-Specific Antibody Representations via Function-Aware Masking
}
\author{Ayan Goel$^{1\dagger}$, Thomas A. Walton$^{2\dagger}$, Amirali Aghazadeh$^{2}$ \vspace{0.2cm} \\  $^1$School of Computer Science, Georgia Institute of Technology\\ $^2$School of Electrical and Computer Engineering, Georgia Institute of Technology\\Correspondence: amiralia@gatech.edu
\\$^\dagger$Equal contribution
}
\date{}
\begin{document}
\maketitle

\vspace{-1cm}
\begin{abstract}
\vspace{-0.6cm}
Antibody-specific language models pretrained via masked language modeling (MLM) learn representations that are critical for downstream sequence design and property prediction tasks. Yet, the corruption process itself is rarely leveraged as a source of inductive bias during pretraining. While preferentially masking complementarity-determining regions (CDRs) improves binding-related predictions, antibodies possess diverse biological priors over a variety of functions. Herein, we introduce function-aware masking, a family of pretraining algorithms that align mask placement with specific functional priors (e.g., from IMGT annotations or structure predictions) to shape the learned representation space. We show that these specialist masking strategies significantly improve performance on their respective objectives, yielding up to a 14\% gain on structure-related tasks and up to a 5.9$\times$ improvement on CDR-related tasks. To further improve performance across multiple functional axes, we develop hybrid masking strategies that integrate multiple priors, balancing reconstruction over binding, structural, and biophysical objectives. Our results demonstrate that informed mask placement provides a parameter-free mechanism for imposing functional inductive biases in antibody language model training.
\end{abstract}

\vspace{-0.7cm}
\section{Introduction}
\label{sec:intro}
\vspace{-0.4cm}
The development of therapeutic antibodies sits at the forefront of drug discovery, providing targeted treatments for complex oncological, autoimmune, and infectious diseases. Learning expressive sequence representations is central to optimizing these biologics for clinical viability. 
Antibody-specific language models trained under the masked language modeling (MLM) objective (abLMs), such as AntiBERTa~\cite{ruffolo2022antiberta, barton2024antiberta2} and AbLang~\cite{olsen2022ablang, olsen2024ablang2}, encode representations that demonstrate strong generalization across diverse sequence design and optimization tasks. 

Improvements to these models have largely stemmed from scaling pretraining data and refining architecture. However, recent work has demonstrated that the corruption process itself serves as an inductive bias that shapes how learned representations generalize across tasks~\cite{ng2025cdr3, Talaei2025}. Specifically, preferential masking of antibody complementarity-determining regions (CDRs) increases transfer performance on CDR infilling, binding affinity, and binding specificity tasks.

Masking of CDRs, however, targets only one facet of antibody function. Therapeutic antibodies must also fold stably, resist aggregation, and exhibit mutational robustness to ensure clinical developability. The residues governing these properties often lie outside of CDRs; they are distributed across the framework region that provides the structural scaffold. The success of CDR masking and its variants points to a broader principle and raises a more general question: can antibody masking strategies be designed using any arbitrary functional prior to align representations with their corresponding downstream tasks?

\begin{figure}[!t]
  \centering
  \includegraphics[width=\linewidth]{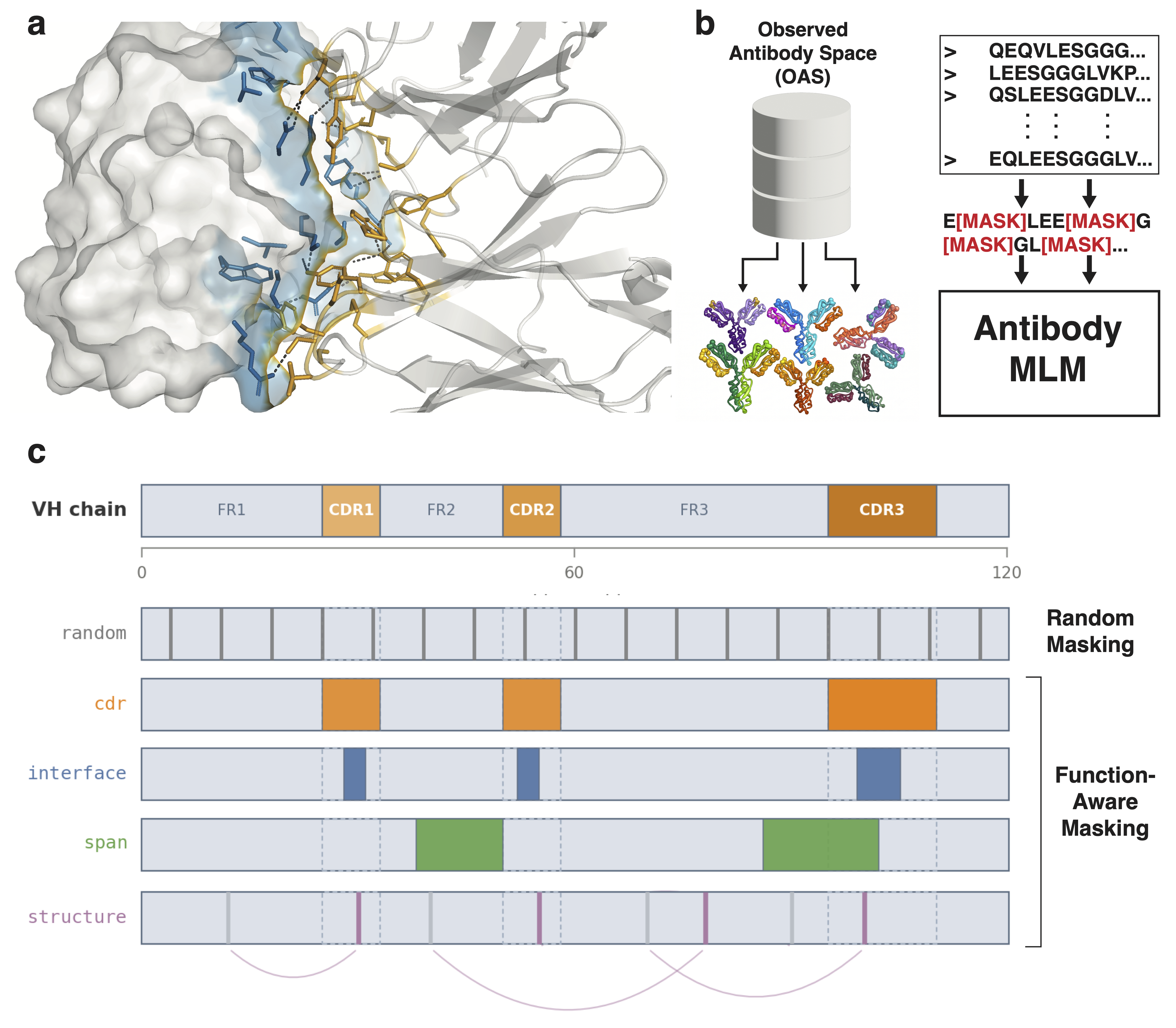}
  \vspace{-0.9cm}
  \caption{{\bf Function-aware masking concentrates representation learning on critical antibody regions.} {\bf a,} The rigid structure of an antibody-antigen complex (PDB ID: 3HFM), illustrating that functional variation is heavily localized to a sparse subset of sequence locations. {\bf b,} Masked language models (MLMs) trained on hundreds of thousands of antibodies implicitly learn how sequence variation affects function through a corruption (masking) and reconstruction objective. {\bf c,} Random masking treats all sequence positions as relevant; however, sequence positions most consequential to downstream fitness are localized to few regions. Function-aware masking leverages functional annotations to guide mask placement, focusing the training signal on these relevant regions.}
  \label{fig:1}
  \vspace{-0.5cm}
\end{figure}

To address this question, we introduce function-aware masking (Figure~\ref{fig:1}), a family of masking strategies that bias mask placement toward the residues most associated with a given functional task. Each strategy scores residues using a functional signal derived from either sequence labels or model predictions (e.g., IMGT labels or IgFold predictions~\cite{ruffolo2023igfold}) and preferentially masks those positions during training (Figure~\ref{fig:1}c). Importantly, function-aware masking shapes only the training objective: resulting models remain sequence-only at inference and incur no additional parameter cost. These masking strategies, hereinafter referred to as specialist strategies, improve performance over random masking on tasks that align with their prior. To consistently drive improvements across multiple functional axes simultaneously, we further introduce hybrid masking strategies that combine functional priors through stochastic mixtures and learning curricula. We evaluate function-aware masking in the pretraining setting and assess performance over five downstream tasks spanning binding, structure, developability, and generative objectives. Our contributions are as follows:

\vspace{-0.2cm}
\begin{enumerate}[leftmargin=14pt]
    \item We introduce function-aware antibody masking, a class of masking algorithms that impose functional inductive biases guided by the downstream task of interest. We open-source our implementation here: \url{https://github.com/amirgroup-codes/function-aware-masking.git}
    \item We demonstrate that specialist masking strategies, focusing on a single biological prior, improve performance over random masking on tasks aligning with their prior. Specialists improved performance on structure-related tasks by up to $14\%$ and CDR-related tasks by up to $5.9\times$.
    \item To further improve generalization, we demonstrate that hybrid masking strategies increase representation quality across multiple functional axes, mitigating performance degradation exhibited by specialists on unaligned tasks.
\end{enumerate}

\vspace{-0.7cm}
\section{Background}
\vspace{-0.3cm}
Corruption is an essential component of MLM training. The most common corruption scheme, replacing tokens with a mask token ([MASK]), has historically been applied uniformly at random across sequence positions with a fixed probability, typically 15\%~\cite{devlin2019bert}. Foundational work in natural language processing demonstrated that mask placement can influence learned representations, priming them for tasks related to inferring semantic structure~\cite{ernie, luke, entitybert}. The extent to which this observation generalizes to proteins is an ongoing question.

\textbf{Biological priors in antibodies.} Antibodies exhibit a highly specialized architecture governed by multiple distinct functional drivers. Antigen recognition is primarily mediated by six CDRs, forming contiguous, highly variable spans. Within these loops, a specific subset of residues (paratopes) forms a direct interface with the target. During affinity maturation, these binding interfaces and their corresponding framework regions undergo somatic hypermutation (SHM), introducing variations relative to the inherited germline sequence. Beyond binding, the overall stable fold of the antibody is maintained by a network of structural contacts, including long-range interactions that form epistatic hubs and dictate conformation. Notably, the aforementioned functional drivers constitute a small proportion of antibody sequences. It follows that much of the corruption applied during MLM training outside of these functionally critical regions is trivially easy to decode given the surrounding context; framework regions in particular are highly conserved across antibodies. 

\textbf{Antibody masking.} Previous works have explored masking strategies for antibodies, primarily focusing on CDRs. Ng \& Briney~\cite{ng2025cdr3} introduced preferential masking of non-templated CDR3, finding that learned representations generalized better to CDR3 residue recovery and improved binding specificity predictions. Talaei et al.~\cite{Talaei2025} extended this work by introducing a hybrid CDR and framework masking strategy, further demonstrating improvements on binding affinity predictions. Whereas previous works focused predominantly on CDRs, this work explores masking strategies for a larger set of function-related tasks aligned with the previously discussed biological priors.

\section{Methods}
\vspace{-0.3cm}
We introduce function-aware masking, a set of masking algorithms leveraging biological priors of antibodies to enable representations across an array of tasks. Function-aware masking is split into two categories: specialist strategies focused primarily on one task, and hybrid strategies that mix priors from many tasks. 

\textbf{Mask formulation and precomputed priors.} To enable targeted mask sampling while limiting computational overhead, we precompute per-residue biological labels and cache them alongside the training data. Labels include CDR locations, paratope probabilities, germline mutation status, and structural topology. During data collation, a mask set $M$ is sampled for sequence $S$ (excluding [CLS]/[SEP] tokens). To isolate the effect of any given strategy compared to the rest, we fix the mask rate at $0.15$. The selected positions then undergo the standard MLM corruption procedure introduced by BERT~\cite{devlin2019bert}: $80\%$ of tokens are replaced by a [MASK] token, $10\%$ by a random amino acid, and the remaining $10\%$ remain unchanged.

\vspace{-0.4cm}
\subsection{Specialist Strategies} 
\label{sec:specialist}
\vspace{-0.3cm}
The goal of specialist masking strategies is to align representations learned by an antibody MLM with a specific downstream task or functional property. For the weight-based strategies, the mask sampler assigns each position a weight $w_i$ encoding its relevance to the target property. Weights are then normalized into per-position masking probabilities calibrated to an expected budget of $15\%$ (Equation~\ref{eq:weighted}). Mask locations are determined with an independent Bernoulli draw (Equation~\ref{eq:sample}). For non-weight-based strategies which depend on previously placed masks (\stratspan{}, \stratstruct{}, \stratstructLR{}), masks are placed iteratively until a fixed budget of 15\% is reached.

\vspace{-0.3cm}
\begin{equation}
    p_i \;=\; \min\!\Bigl(0.15\,\tfrac{w_i}{\bar w},\, 1\Bigr)\,\mathbf{1}[i\in S],
    \qquad \bar w = \tfrac{1}{|S|}\!\sum_{j\in S} w_j,
    \qquad \mathbb{E}[|M|] = 0.15\,|S|.
    \label{eq:weighted}
  \end{equation}
\begin{equation}
    m_i \sim \mathrm{Bernoulli}(p_i) \ \text{ independently},
    \qquad M = \{\, i : m_i = 1 \,\}.
    \label{eq:sample}
\end{equation}

We evaluate seven specialist strategies designed to target distinct functional, evolutionary, and structural priors. Details on exact implementations can be found in Section~\ref{sec:experiments}.

\textbf{\stratcdr{}}: Targets hypervariable regions by over-sampling the complementarity-determining loops ($w_\text{FW}\!:\!w_\text{CDR1}\!:\!w_\text{CDR2}\!:\!w_\text{CDR3}\!=\!1\!:\!3\!:\!3\!:\!6$), generalizing CDR3 masking to the remaining CDRs~\cite{ng2025cdr3}.

\textbf{\stratspan{}}: Topology-agnostic, masks contiguous spans of sequence positions using a geometric distribution ($p\!=\!0.2$, $\ell_\text{max}\!=\!10$).

\textbf{\stratint{}}: Targets the physical binding interface by placing a $6:1$ weight ratio on paratope-contacting residues.

\textbf{\stratgerm{}}: Targets the evolutionary trajectory of the antibody by placing a $6:1$ weight ratio on germline-mutated residues.

\textbf{\stratinter{}}: A joint specialist that multiplies the paratope and germline weights: $w_i = w^{\text{para}}_i \cdot w^{\text{germ}}_i$. This strategy targets residues responsible for affinity maturation, concentrating masks on positions that are both antigen-contacting and somatically mutated from the germline.

\textbf{\stratstruct{}}: A 3D-aware scheme that weights residues based on predicted spatial proximity. Utilizes the structure's five nearest neighbors to mask structural neighborhoods. Masks are sampled across neighborhoods, leaving enough spatial context to decode the masked residues.

\textbf{\stratstructLR{}}: An extension of \stratstruct{} that applies a stricter partitioning rule: only structural couplings greater than four sequence indices away are considered. Prioritizes modeling long-range contacts during reconstruction.

\vspace{-0.4cm}
\subsection{Hybrid Strategies}
\label{sec:hybrid}
\vspace{-0.2cm}
Specialist strategies target one desired antibody property; however, there may be many such properties to optimize. To resolve this, we develop hybrid strategies that stochastically mix functional priors over the course of training. Given a set of $K$ masking strategies, we define the mixing distribution $\pi(t) = (p_1(t), \dots, p_K(t))$, where $p_i(t)$ represents the probability of drawing specialist strategy $i$ at training step $t$ ($\sum_s p_s(t)=1$). Mixing distributions follow two principles: masks should be sampled uniformly before specializing~\cite{yang-etal-2023-learning}, and the best performing specialists should have the highest probability of being sampled. Sampling is performed by drawing strategy $i \sim \pi(t)$ and applying masks per sequence. We dynamically adjust $\pi(t)$ over the course of training to bias which functional priors are drawn. Details on mixing distribution construction are available in Appendix~\ref{ap:hybrid_masking}.

Hybrid strategies interpolate between the following specialists unless specified otherwise, referenced in the following order: [\stratuniform{}, \stratcdr{}, \stratspan{}, \stratstructLR{}, \stratint{}, \stratgerm{}]. We evaluate the following hybrid strategies:

\textbf{\strathyb{}:} Basis for the following hybrid strategies. Emphasizes \stratuniform{} and \stratspan{} masking before gradually shifting towards specialists.

\textbf{\strathybrand{}:} A randomized hybrid control strategy which samples from $\pi(t)$ uniformly.

\textbf{\strathybst{}:} Skews the mixing distribution heavily toward \stratuniform{} and \stratspan{} to start. Ends with the same distribution as \strathyb{}. Delays transitions in mixing across more time steps. 

\textbf{\strathybrv{}:} Reverses the direction in which \strathybst{} is applied.

\textbf{\strathybw{}:} Starts with a general strategy similar to \strathybst{}, shifts focus to \stratint{} masking, followed by a mix of \stratgerm{} and \stratint{} masking, before finally ending with a more balanced distribution.

\textbf{\strathybpb{}:} Identical schedule to \strathyb{}, but draws a single specialist per batch of sequences instead of per sequence. 

\textbf{\strathybws{}:} Continued pretraining instead of a curriculum. Takes a fully trained \stratint{} masking model and trains for an additional 50,000 steps on a balanced mixing distribution.

\vspace{-0.4cm}
\section{Experiments}
\label{sec:experiments}
\vspace{-0.2cm}
We study function-aware masking in the pretraining setting, training models parameterized by the RoFormer architecture~\cite{su2024roformer}. Specifically, we instantiate a medium-sized RoFormer model (38.1M parameters) utilizing the AntiBERTa2 tokenizer \citep{barton2024antiberta2}, AdamW optimizer (peak LR $5\!\times\!10^{-5}$, 5\,\% warmup, cosine decay), batch size 64, fp16, and 125,000 training steps. All masking strategies are evaluated using this setup with the exception of \strathybws{}, which trains for an additional 50,000 steps on top of the \stratint{} model checkpoint at a reduced learning rate ($2\!\times\!10^{-5}$). Each model is trained three times across different initialization seeds. Pretraining runs take approximately five hours on one RTX A6000 under fp16; \strathybws{} adds 2 hours per seed. One time IgFold structure prediction required $\approx$240 GPU hours.

\textbf{Data.} We sampled 500,000 heavy-chain variable-domain (VH) sequences from the Observed Antibody Space~\cite{kovaltsuk2018oas} and filtered them to the 20 canonical amino acids and lengths in $[80,160]$, leaving 497{,}309 total sequences. Models are pretrained on the same 90/10 training and hold-out split across each run. CDR1/2/3 boundaries are taken from the OAS IMGT annotations, with ANARCI~\cite{dunbar2014anarci} as a fallback for unannotated sequences. Per-residue paratope probabilities are predicted by a teacher model: AntiBERTa2 fine-tuned with a per-token classification head. The classification head predicts over the TDC SAbDab\_Liberis paratope set~\cite{liberis2018parapred,dunbar2014sabdab}, storing residue labels in $[0,1]$. 
Germline labels are computed against a per-gene consensus built from the training corpus; sequences are grouped by V- and J-gene calls, with position-wise majority voting forming a consensus for genes represented by at least 20 sequences. A residue is labeled mutated ($1.0$) if it differs from the consensus, germline ($0.0$) if it matches, or CDR3 junction ($0.5$) where no consensus is available. 
Structures are predicted with IgFold~\cite{ruffolo2023igfold} (experimental X-ray structures exist for $<1\,\%$ of OAS), requiring approximately 240 GPU-hours. We build a per-residue $k$-nearest neighbors graph using $C_\alpha-C_\alpha$ distances; structure-aware strategies mask residues within these neighborhoods ($k=5$). \stratstructLR{} masking follows this logic but restricts neighborhoods to residues more than four indices apart in sequence, targeting long-range interactions. Datasets used for paratope and structure prediction are presplit to ensure no leakage with the evaluation sequences.

\textbf{Evaluation.} We evaluate learned representations in two stages: first, the model is pretrained with a given masking strategy; then, the pretrained encoder is frozen and embeddings are extracted from the last layer as input to downstream probes. We assess seven metrics to isolate distinct functional properties captured by learned representations: zero-shot CDR3 infilling on held-out OAS heavy chains (CDR3); paratope AUPRC (Para) and MCC on TDC SAbDab\_Liberis; contact map AUROC (Cont) and long-range precision-at-L on SAbDab; structure-probe Spearman $\rho$ on AB-Bind and SAbDab (Struct); and developability macro-Spearman $\rho$ on TDC TAP (Dev). All metrics other than CDR3 infilling are computed with linear or bilinear probes. Probe architectures, optimization, and data splits are provided in Appendix~\ref{ap:experiments}.

\begin{figure}[!h]
  \centering
  \includegraphics[width=\linewidth]{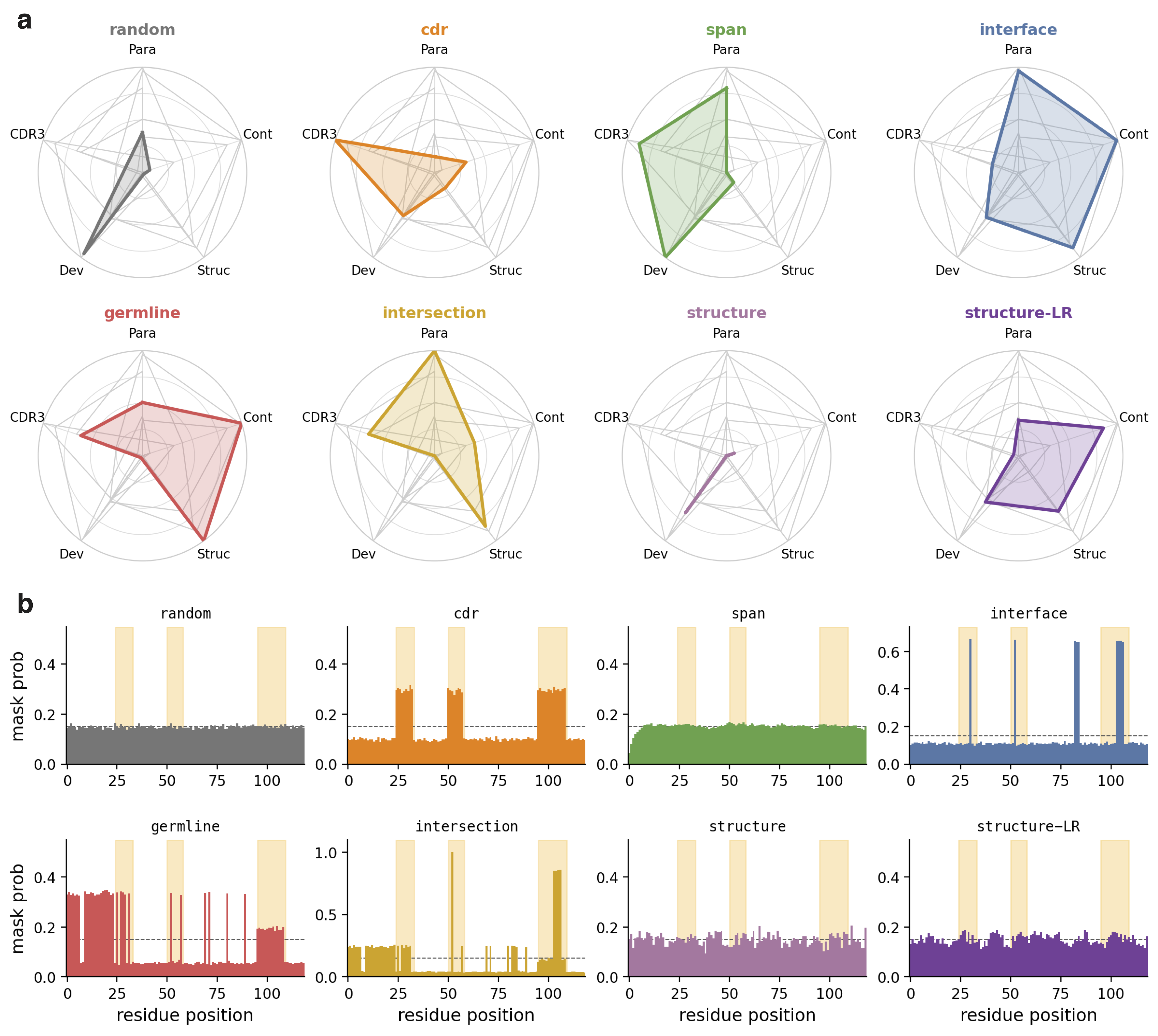}
  \caption{{\bf Mask placement distribution influences downstream task performance in antibody MLMs. a,} Performance profiles of eight distinct masking strategies evaluated across predictive tasks spanning five functional axes: paratope prediction (Para), contact map prediction (Cont), structure probe (Struc), developability (Dev), and CDR3 infilling (CDR3). Each strategy excels at a distinct subset of tasks, improving over random masking in at least one area. {\bf b,} Empirical probability distribution of mask placements across sequence positions. Comparing these distributions to performance in {\bf a} demonstrates that strategies excel at tasks functionally aligned with their masking locations.}
  \label{fig:2}
  \vspace{-0.5cm}
\end{figure}

\vspace{-0.5cm}
\section{Results}
\label{sec:results}
\vspace{-0.4cm}

\textbf{Function-aware masking aligns learned representations with biological priors.} As detailed in Figure~\ref{fig:2} and Table~\ref{tab:results_specialist}, embedding biological priors into the MLM corruption process yields representations that successfully specialize in their targeted functional tasks. With the exception of structure-based strategies, every specialist masking strategy achieved the highest score on at least one metric.

\begin{table}[h]
    \vspace{-0.4cm}
    \caption{Specialist masking strategy results. Metrics are defined in Section~\ref{sec:experiments} and strategies
  are defined in Section~\ref{sec:specialist}. \stratuntrained{} is a randomly initialized control (no
  pretraining). Higher is better on all reported metrics.}
    \label{tab:results_specialist}
    \centering
    \footnotesize
    \resizebox{\textwidth}{!}{%
    \begin{tabular}{lccccccc}
    \toprule
    Strategy & CDR3 & P.~AUPRC & P.~MCC & C.~AUROC & C.~P@L & Str.~$\rho$ & Dev.~$\rho$ \\
    \midrule
    \stratuntrained{}    & 0.000$_{\pm 0.000}$ & 0.130$_{\pm 0.001}$ & 0.164$_{\pm 0.001}$ & 0.591$_{\pm 0.004}$ & 0.071$_{\pm 0.004}$ & 0.087$_{\pm 0.002}$ & 0.102$_{\pm 0.021}$ \\
    \midrule
    \stratuniform{}      & 0.039$_{\pm 0.011}$ & 0.821$_{\pm 0.012}$ & 0.626$_{\pm 0.002}$ & 0.970$_{\pm 0.003}$ & 0.604$_{\pm 0.011}$ & 0.592$_{\pm 0.023}$ & 0.358$_{\pm 0.022}$ \\
    \stratcdr{}          & \textbf{0.231}$_{\pm 0.005}$ & 0.805$_{\pm 0.025}$ & 0.636$_{\pm 0.021}$ & 0.973$_{\pm 0.009}$ & 0.627$_{\pm 0.079}$ & 0.597$_{\pm 0.012}$ & 0.316$_{\pm 0.018}$ \\
    \stratspan{}         & 0.204$_{\pm 0.006}$ & 0.852$_{\pm 0.003}$ & 0.637$_{\pm 0.013}$ & 0.968$_{\pm 0.006}$ & 0.597$_{\pm 0.041}$ & 0.595$_{\pm 0.019}$ & \textbf{0.362}$_{\pm 0.020}$ \\
    \stratstruct{}       & 0.000$_{\pm 0.001}$ & 0.794$_{\pm 0.014}$ & 0.607$_{\pm 0.024}$ & 0.969$_{\pm 0.005}$ & 0.605$_{\pm 0.044}$ & 0.591$_{\pm 0.010}$ & 0.331$_{\pm 0.037}$ \\
    \stratstructLR{}     & 0.012$_{\pm 0.007}$ & 0.818$_{\pm 0.021}$ & 0.625$_{\pm 0.017}$ & 0.978$_{\pm 0.003}$ & 0.678$_{\pm 0.020}$ & 0.612$_{\pm 0.009}$ & 0.319$_{\pm 0.047}$ \\
    \stratint{}          & 0.061$_{\pm 0.034}$ & 0.864$_{\pm 0.011}$ & \textbf{0.645}$_{\pm 0.017}$ & 0.979$_{\pm 0.003}$ & 0.691$_{\pm 0.031}$ & 0.620$_{\pm 0.012}$ & 0.318$_{\pm 0.047}$ \\
    \stratgerm{}         & 0.144$_{\pm 0.012}$ & 0.831$_{\pm 0.014}$ & 0.615$_{\pm 0.030}$ & \textbf{0.980}$_{\pm 0.005}$ & \textbf{0.692}$_{\pm 0.056}$ & \textbf{0.624}$_{\pm 0.016}$ & 0.270$_{\pm 0.136}$ \\
    \stratinter{}        & 0.154$_{\pm 0.003}$ & \textbf{0.866}$_{\pm 0.006}$ & 0.644$_{\pm 0.035}$ & 0.974$_{\pm 0.000}$ & 0.635$_{\pm 0.002}$ & 0.618$_{\pm 0.008}$ & 0.267$_{\pm 0.033}$ \\
    \bottomrule
  \end{tabular}}
  \vspace{-0.3cm}
\end{table}

By directing the masking objective toward residues critical for specific functions, these strategies effectively align the learned representations with their corresponding downstream tasks. For example, \stratcdr{} masking performed best on CDR3 infilling, corroborating previously reported observations~\cite{ng2025cdr3, Talaei2025}. The \stratint{} and \stratgerm{} masking strategies performed the best overall, scoring highest across multiple structure- and paratope-related tasks. Interestingly, while \stratinter{} masking improved representations for paratope- and CDR-related tasks, it largely failed to retain developability and structural gains from the \stratint{} and \stratgerm{} strategies it was derived from. This performance gap likely arises due to \stratinter{} masking concentrating masks in fewer sequence locations, indicating that maintaining broad competency requires a sufficient diversity of mask locations across the entire sequence.

\stratstruct{} and \stratstructLR{} masking performed worse relative to the other specialists. The residues most affiliated with modulating structural functions lie in the CDRs, which are targeted more frequently with \stratint{} and \stratinter{} masking; modeling distal contacts in framework regions provides little additional training signal.
 
While \stratspan{} masking does not incorporate a specific biological prior, masking contiguous regions increases the difficulty of the MLM task, a mechanism associated with learning better representations in natural language processing~\cite{joshi2020spanbert}. Indeed, \stratspan{} masking generalized well across paratope, CDR3, and developability tasks. However, \stratspan{} masking fell short on tasks that \stratint{} and \stratgerm{} masking specialized in, indicating that the placement of masks is more important than the shape for these tasks. 

Overall, by targeting residues most associated with their biological prior, specialist masking strategies effectively align learned representations with their intended downstream tasks. Furthermore, with the exception of \stratstruct{} masking, every specialist also improved over \stratuniform{} masking on multiple metrics, including \stratint{} and \stratgerm{} masking, which demonstrated gains across nearly all functional axes. These results indicate both the upside and necessity of moving away from random masking in antibody MLMs.

\textbf{Hybrid masking strategies balance generalization across functional tasks.} While specialist strategies excelled on their targeted objectives, their performance can degrade on unaligned tasks. To achieve more robust representations, we evaluated seven hybrid masking strategies, each designed to combine multiple biological priors via stochastic mixtures and training curricula. 

Table~\ref{tab:results_hybrid} details the performance of each hybrid method. The best performing hybrid strategies were \strathybws{} and \strathybrv{} masking. Interestingly, \strathybrv{} masking performed best on three of seven tasks despite contradicting the design principle of the other hybrids: masks should be placed uniformly before specializing. This indicates that important functional axes may be learned early during pretraining, after which specializing provides diminishing returns.

\begin{figure}[!t]
  \centering
  \includegraphics[width=\linewidth]{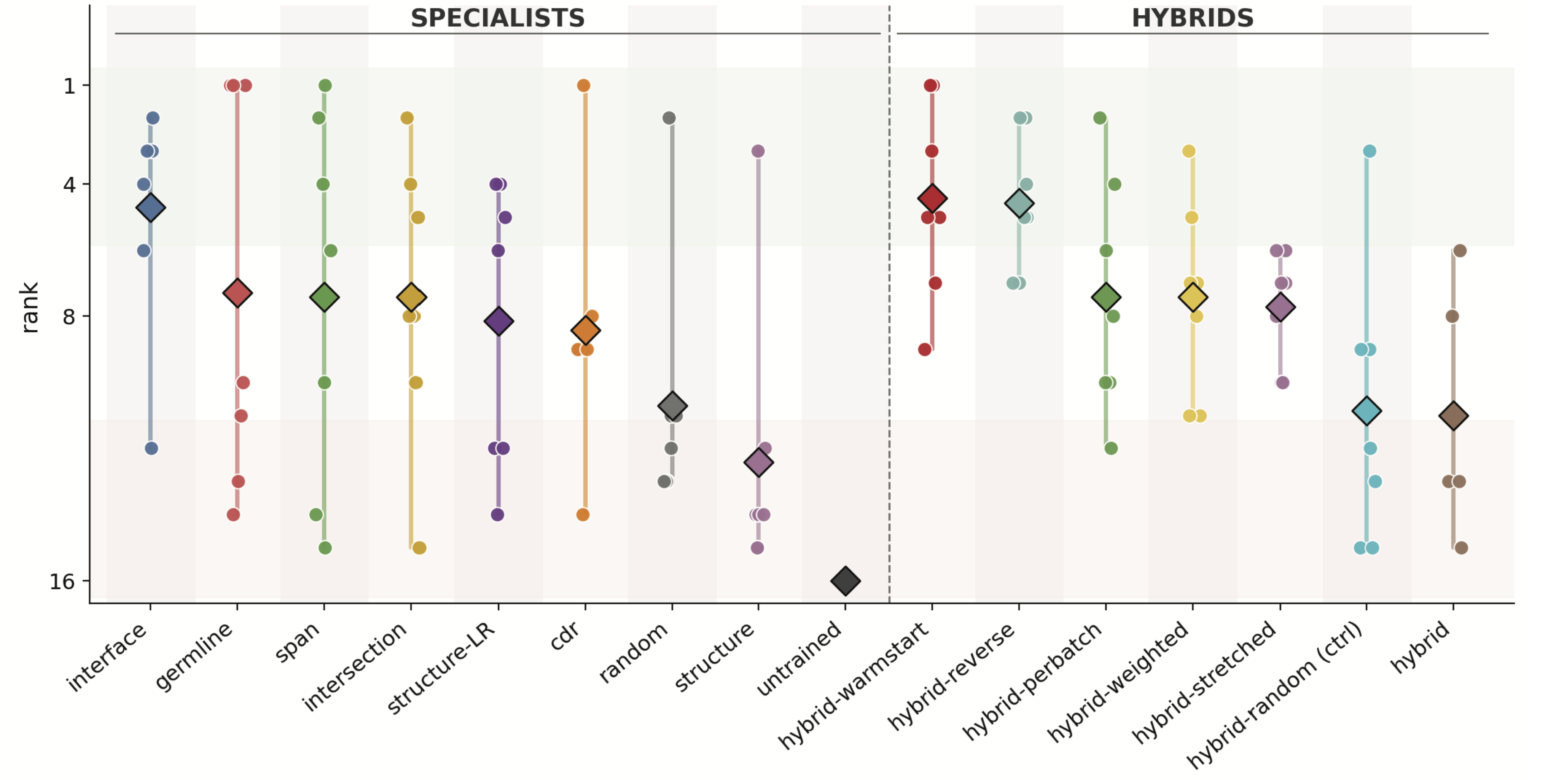}
  \vspace{-0.8cm}
  \caption{{\bf Hybrid masking schemes mitigate task-specific trade-offs.} Performance rankings of specialist versus hybrid masking strategies evaluated across multiple tasks. While specialist strategies can achieve high-ranking performance on specific tasks, they exhibit significant performance drops on others. Hybrid strategies mitigate this behavior by combining specialist strategies, compensating for individual task inadequacies to achieve a higher average ranking.}
  \label{fig:3}
  \vspace{-0.4cm}
\end{figure}

\strathyb{} masking performed the worst overall, being superseded in performance by both the \stratuniform{} and \strathybrand{} masking controls. While \strathybrand{} masking outperformed \strathyb{} masking by uniformly sampling from specialists, the remaining hybrid strategies surpassed both controls. This demonstrates that the choice of mixing distribution and learning curriculum is essential for learning representations that generalize across functional axes. These strategies achieved a higher average task rank (Figure~\ref{fig:3}), particularly on structure-based tasks. Alongside their competitive average task rank compared to their specialist counterparts, hybrid strategies also demonstrated a lower overall task variance. While hybrid strategies seldom performed best on any individual task, their consistent performance across all functional axes indicates their ability to successfully bridge the gap between narrow specialization and broad generalization.

\begin{table}[!htbp]
    \vspace{-0.3cm}
    \caption{Hybrid strategy results. Metrics are defined in Section~\ref{sec:experiments} and strategies are
  defined in Section~\ref{sec:hybrid}. Higher is better on all reported metrics. Each strategy underwent the
  same training setup with the exception of \strathybws{}, which was trained for an additional 50,000 steps.}
    \label{tab:results_hybrid}
    \centering
    \footnotesize
    \resizebox{\textwidth}{!}{%
    \begin{tabular}{lccccccc}
    \toprule
    Strategy & CDR3 & P.~AUPRC & P.~MCC & C.~AUROC & C.~P@L & Str.~$\rho$ & Dev.~$\rho$ \\
    \midrule
    \strathybrand{}      & \textbf{0.201}$_{\pm 0.010}$ & 0.833$_{\pm 0.026}$ & 0.629$_{\pm 0.007}$ & 0.969$_{\pm 0.003}$ & 0.593$_{\pm 0.022}$ & 0.589$_{\pm 0.016}$ & 0.292$_{\pm 0.070}$ \\
    \strathyb{}          & 0.198$_{\pm 0.002}$ & 0.814$_{\pm 0.017}$ & 0.603$_{\pm 0.018}$ & 0.973$_{\pm 0.006}$ & 0.620$_{\pm 0.056}$ & 0.597$_{\pm 0.017}$ & 0.289$_{\pm 0.005}$ \\
    \strathybst{}        & 0.196$_{\pm 0.008}$ & 0.834$_{\pm 0.019}$ & 0.628$_{\pm 0.041}$ & 0.977$_{\pm 0.003}$ & 0.658$_{\pm 0.033}$ & 0.604$_{\pm 0.018}$ & 0.304$_{\pm 0.060}$ \\
    \strathybrv{}        & 0.199$_{\pm 0.007}$ & 0.842$_{\pm 0.017}$ & 0.636$_{\pm 0.020}$ & \textbf{0.980}$_{\pm 0.005}$ & \textbf{0.691}$_{\pm 0.045}$ & 0.615$_{\pm 0.011}$ & \textbf{0.329}$_{\pm 0.029}$ \\
    \strathybpb{}        & 0.200$_{\pm 0.002}$ & 0.848$_{\pm 0.024}$ & 0.660$_{\pm 0.001}$ & 0.973$_{\pm 0.004}$ & 0.626$_{\pm 0.030}$ & 0.593$_{\pm 0.020}$ & 0.317$_{\pm 0.016}$ \\
    \strathybw{}         & 0.195$_{\pm 0.005}$ & 0.850$_{\pm 0.011}$ & 0.648$_{\pm 0.034}$ & 0.974$_{\pm 0.007}$ & 0.644$_{\pm 0.048}$ & 0.593$_{\pm 0.017}$ & 0.299$_{\pm 0.040}$ \\
    \strathybws{}        & 0.168$_{\pm 0.003}$ & \textbf{0.868}$_{\pm 0.009}$ & \textbf{0.672}$_{\pm 0.017}$ & 0.977$_{\pm 0.004}$ & 0.666$_{\pm 0.030}$ & \textbf{0.619}$_{\pm 0.012}$ & 0.317$_{\pm 0.051}$ \\
    \bottomrule
  \end{tabular}}
\end{table}
\vspace{-0.5cm}

\section{Conclusion}
\label{sec:conclusion}
\vspace{-0.4cm}
In this work, we introduced function-aware masking, a framework that leverages biological priors to impose targeted inductive biases during the pretraining of antibody language models. While standard random masking treats all residues equally, function-aware masking aligns the corruption objective directly with the diverse requirements of therapeutic drug discovery. We demonstrated that specialist masking strategies improve downstream performance on their aligned tasks, achieving up to a $14\%$ improvement on structure-related tasks and up to a $5.9\times$ gain on CDR-related tasks, all without incurring additional parameter costs at inference.

As antibody optimization requires balancing multiple complex properties simultaneously, we further developed hybrid masking strategies to mitigate the potential trade-offs of narrow specialization. By combining multiple functional priors via stochastic mixtures and training curricula, hybrid models successfully maintain broad generalization and consistently outperform random masking across a diverse array of functional tasks. Ultimately, our findings reinforce that mask placement acts as an inductive bias over learned representations, offering a flexible, parameter-free method for training antibody language models.

\textbf{Limitations.} Function-aware masking requires per-residue labels, which may vary in accuracy or acquisition complexity depending on the strategy. Evaluations are done on a medium-sized RoFormer model ($\approx$38M params); further testing would elucidate the impact of masking priors on a larger scale. Hybrid strategies were manually designed; systematic optimization of mixing distributions may yield additional gains.

\clearpage
\newpage

\textbf{Acknowledgments.} This work was supported in part by the HIVES program at Georgia Tech Research Institute (GTRI) and Georgia Institute of Technology start-up funds.

\bibliographystyle{ieeetr}
\bibliography{references}

\newpage

\appendix
\section*{Appendix}

\textbf{Future work.} Masking strategies represent a promising mechanism for shaping the organization of learned representations during pretraining, yet the extent to which they influence downstream generalization remains poorly understood. Organizing representation space by therapeutic properties remains a central problem in protein engineering~\cite{tsui2024recoveringhigherorderinteractionsprotein, tsui2025sparseautoencoderslownprotein} and variant effect prediction~\cite{golf-walton25a}. An interesting future direction could incorporate explainability-driven approaches into the masking strategy~\cite{tsui2025shapzero}, enabling interpretable control over corruption schemes.

\section{Additional Results}
We present the full results table in this section, combining experiments from Table~\ref{tab:results_specialist} and Table~\ref{tab:results_hybrid}.

\begin{table}[!h]
  \caption{Full results table across all strategies presented in the main text.}
  \label{tab:results-master}
  \centering
  \footnotesize
  \resizebox{\textwidth}{!}{%
  \begin{tabular}{lccccccc}
    \toprule
    Strategy & CDR3 & P.~AUPRC & P.~MCC & C.~AUROC & C.~P@L & Str.~$\rho$ & Dev.~$\rho$ \\
    \midrule
    \stratuntrained{}    & 0.000$_{\pm 0.000}$ & 0.130$_{\pm 0.001}$ & 0.164$_{\pm 0.001}$ & 0.591$_{\pm 0.004}$ & 0.071$_{\pm 0.004}$ & 0.087$_{\pm 0.002}$ & 0.102$_{\pm 0.021}$ \\
    \midrule
    \stratuniform{}      & 0.039$_{\pm 0.011}$ & 0.821$_{\pm 0.012}$ & 0.626$_{\pm 0.002}$ & 0.970$_{\pm 0.003}$ & 0.604$_{\pm 0.011}$ & 0.592$_{\pm 0.023}$ & 0.358$_{\pm 0.022}$ \\
    \stratcdr{}          & 0.231$_{\pm 0.005}$ & 0.805$_{\pm 0.025}$ & 0.636$_{\pm 0.021}$ & 0.973$_{\pm 0.009}$ & 0.627$_{\pm 0.079}$ & 0.597$_{\pm 0.012}$ & 0.316$_{\pm 0.018}$ \\
    \stratspan{}         & 0.204$_{\pm 0.006}$ & 0.852$_{\pm 0.003}$ & 0.637$_{\pm 0.013}$ & 0.968$_{\pm 0.006}$ & 0.597$_{\pm 0.041}$ & 0.595$_{\pm 0.019}$ & 0.362$_{\pm 0.020}$ \\
    \stratstruct{}       & 0.000$_{\pm 0.001}$ & 0.794$_{\pm 0.014}$ & 0.607$_{\pm 0.024}$ & 0.969$_{\pm 0.005}$ & 0.605$_{\pm 0.044}$ & 0.591$_{\pm 0.010}$ & 0.331$_{\pm 0.037}$ \\
    \stratstructLR{}     & 0.012$_{\pm 0.007}$ & 0.818$_{\pm 0.021}$ & 0.625$_{\pm 0.017}$ & 0.978$_{\pm 0.003}$ & 0.678$_{\pm 0.020}$ & 0.612$_{\pm 0.009}$ & 0.319$_{\pm 0.047}$ \\
    \stratint{}          & 0.061$_{\pm 0.034}$ & 0.864$_{\pm 0.011}$ & 0.645$_{\pm 0.017}$ & 0.979$_{\pm 0.003}$ & 0.691$_{\pm 0.031}$ & 0.620$_{\pm 0.012}$ & 0.318$_{\pm 0.047}$ \\
    \stratgerm{}         & 0.144$_{\pm 0.012}$ & 0.831$_{\pm 0.014}$ & 0.615$_{\pm 0.030}$ & 0.980$_{\pm 0.005}$ & 0.692$_{\pm 0.056}$ & 0.624$_{\pm 0.016}$ & 0.270$_{\pm 0.136}$ \\
    \stratinter{}        & 0.154$_{\pm 0.003}$ & 0.866$_{\pm 0.006}$ & 0.644$_{\pm 0.035}$ & 0.974$_{\pm 0.000}$ & 0.635$_{\pm 0.002}$ & 0.618$_{\pm 0.008}$ & 0.267$_{\pm 0.033}$ \\
    \strathybrand{}      & 0.201$_{\pm 0.010}$ & 0.833$_{\pm 0.026}$ & 0.629$_{\pm 0.007}$ & 0.969$_{\pm 0.003}$ & 0.593$_{\pm 0.022}$ & 0.589$_{\pm 0.016}$ & 0.292$_{\pm 0.070}$ \\
    \strathyb{}          & 0.198$_{\pm 0.002}$ & 0.814$_{\pm 0.017}$ & 0.603$_{\pm 0.018}$ & 0.973$_{\pm 0.006}$ & 0.620$_{\pm 0.056}$ & 0.597$_{\pm 0.017}$ & 0.289$_{\pm 0.005}$ \\
    \strathybst{}        & 0.196$_{\pm 0.008}$ & 0.834$_{\pm 0.019}$ & 0.628$_{\pm 0.041}$ & 0.977$_{\pm 0.003}$ & 0.658$_{\pm 0.033}$ & 0.604$_{\pm 0.018}$ & 0.304$_{\pm 0.060}$ \\
    \strathybrv{}        & 0.199$_{\pm 0.007}$ & 0.842$_{\pm 0.017}$ & 0.636$_{\pm 0.020}$ & 0.980$_{\pm 0.005}$ & 0.691$_{\pm 0.045}$ & 0.615$_{\pm 0.011}$ & 0.329$_{\pm 0.029}$ \\
    \strathybpb{}        & 0.200$_{\pm 0.002}$ & 0.848$_{\pm 0.024}$ & 0.660$_{\pm 0.001}$ & 0.973$_{\pm 0.004}$ & 0.626$_{\pm 0.030}$ & 0.593$_{\pm 0.020}$ & 0.317$_{\pm 0.016}$ \\
    \strathybw{}         & 0.195$_{\pm 0.005}$ & 0.850$_{\pm 0.011}$ & 0.648$_{\pm 0.034}$ & 0.974$_{\pm 0.007}$ & 0.644$_{\pm 0.048}$ & 0.593$_{\pm 0.017}$ & 0.299$_{\pm 0.040}$ \\
    \strathybws{}        & 0.168$_{\pm 0.003}$ & 0.868$_{\pm 0.009}$ & 0.672$_{\pm 0.017}$ & 0.977$_{\pm 0.004}$ & 0.666$_{\pm 0.030}$ & 0.619$_{\pm 0.012}$ & 0.317$_{\pm 0.051}$ \\
    \bottomrule
  \end{tabular}}
\end{table}

\section{Implementation Details}
\begin{figure}[!h]
  \centering
  \includegraphics[width=\linewidth]{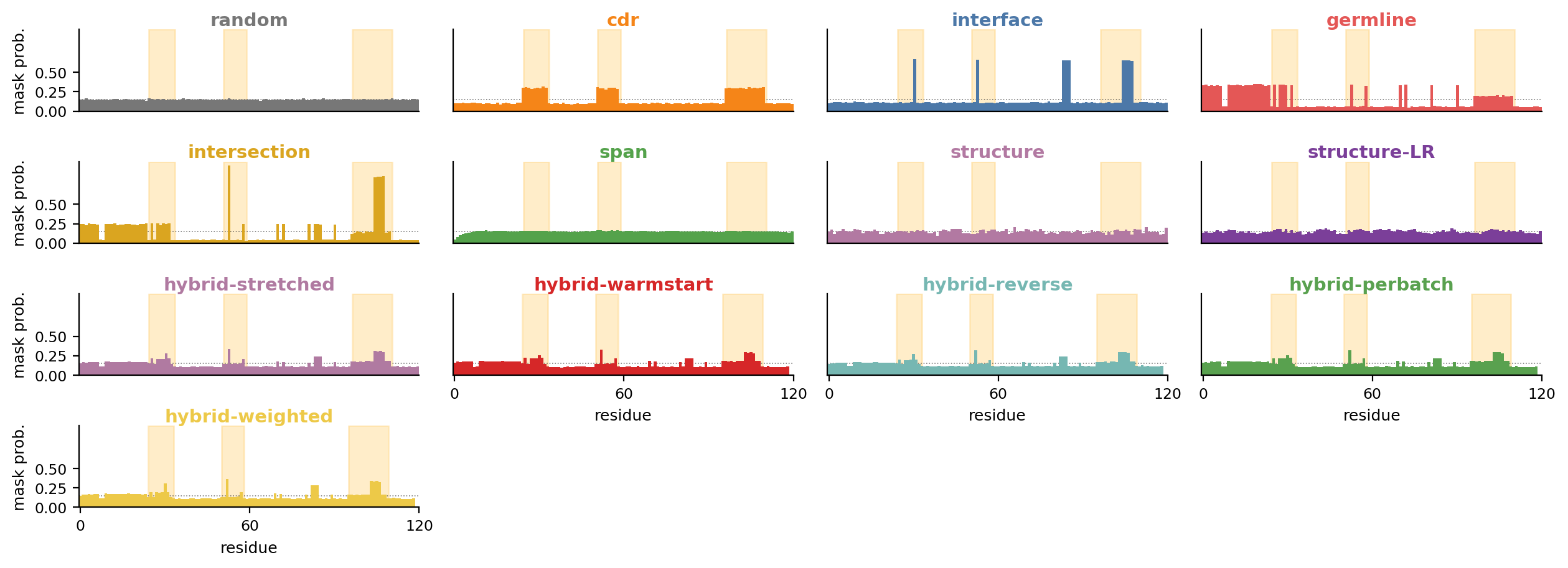}
  \caption{Empirical per-position mask probabilities for all thirteen strategies evaluated in this work.}
  \label{ap:masking_schemes}
\end{figure}

This section outlines implementation details for specialist and hybrid masking schemes presented in the main text. Furthermore, we visualize empirical probability distributions in Figure~\ref{ap:masking_schemes} for thirteen methods tested in this work (excluding \strathyb{} and \strathybrand{} masking) with CDRs highlighted for reference. 

\subsection{Specialists}
\label{ap:specialist_masking}

\textbf{Structure and structure-LR.} Structures for each antibody are predicted using IgFold~\cite{ruffolo2023igfold}, an antibody-specific structure predictor. We build a per-residue $k$-nearest neighbors graph by computing Euclidean $C_{\alpha}-C_{\alpha}$ distance. \stratstruct{} utilizes $k=5$ nearest neighbors, masking residues which form structural neighborhoods. From these neighborhoods, only a few masks are sampled. This is done to provide enough structural context to decode the masked residue. Structure-LR follows the same logic, but restricts residue neighborhoods to only include corresponding sequence locations given that they are more than four indices apart.

\subsection{Hybrid Masking Strategies}
\label{ap:hybrid_masking}
Hybrid masking strategies sample masks from a mixing distribution specified by a timestep $t$, combining specialist strategies as a stochastic mixture. We describe these learning curricula in this section, indicating the mixing distribution for each training step interval.

\textbf{Hybrid (default).} The baseline curriculum for the remaining hybrid strategies. Starts with higher emphasis on \stratuniform{} and \stratspan{} masking, gradually transitioning to more weight on \stratcdr{} and \stratint{}. After step 40,000, the mixing strategy covers each strategy more generally. 

\begin{table}[htbp]
    \centering
    \caption{Mixing schedule for hybrid (default).}
    \label{tab:sched_default}
    \footnotesize
    \begin{tabular}{lcccccc}
        \toprule
        step & random & cdr & span & structure & interface & germline \\
        \midrule
        0     & 0.30 & 0.15 & 0.30 & 0.10 & 0.10 & 0.05 \\
        6250  & 0.10 & 0.30 & 0.10 & 0.15 & 0.25 & 0.10 \\
        18750 & 0.10 & 0.20 & 0.10 & 0.15 & 0.20 & 0.25 \\
        40000 & 0.10 & 0.20 & 0.10 & 0.15 & 0.20 & 0.25 \\
        \bottomrule
    \end{tabular}
\end{table}

\textbf{Hybrid-stretched.} Begins with more probability mass assigned to \stratuniform{} and \stratspan{}, and gradually stretches out the same strategy as \strathyb{} over more steps.

\begin{table}[htbp]
    \centering
    \caption{Mixing schedule for hybrid-stretched.}
    \label{tab:sched_stretched}
    \footnotesize
    \begin{tabular}{lcccccc}
        \toprule
        step & random & cdr & span & structure & interface & germline \\
        \midrule
        0     & 0.50 & 0.05 & 0.40 & 0.00 & 0.05 & 0.00 \\
        20000 & 0.10 & 0.30 & 0.10 & 0.10 & 0.30 & 0.10 \\
        50000 & 0.05 & 0.15 & 0.05 & 0.15 & 0.30 & 0.30 \\
        90000 & 0.10 & 0.20 & 0.10 & 0.15 & 0.20 & 0.25 \\
        \bottomrule
    \end{tabular}
\end{table}

\textbf{Hybrid-reverse.} Same strategy as \strathybst{} but in reverse, stretched over more steps.

\begin{table}[htbp]
    \centering
    \caption{Mixing schedule for hybrid-reverse.}
    \label{tab:sched_reverse}
    \footnotesize
    \begin{tabular}{lcccccc}
        \toprule
        step & random & cdr & span & structure & interface & germline \\
        \midrule
        0      & 0.05 & 0.10 & 0.05 & 0.15 & 0.30 & 0.35 \\
        35000  & 0.05 & 0.15 & 0.05 & 0.15 & 0.30 & 0.30 \\
        75000  & 0.10 & 0.30 & 0.10 & 0.10 & 0.30 & 0.10 \\
        105000 & 0.50 & 0.05 & 0.40 & 0.00 & 0.05 & 0.00 \\
        \bottomrule
    \end{tabular}
\end{table}

\textbf{Hybrid-weighted.} Focuses more heavily on the best performing specialist, \stratint{}.

\begin{table}[!h]
    \centering
    \caption{Mixing schedule for hybrid-weighted.}
    \label{tab:sched_weighted}
    \footnotesize
    \begin{tabular}{lcccccc}
        \toprule
        step & random & cdr & span & structure & interface & germline \\
        \midrule
        0     & 0.40 & 0.05 & 0.45 & 0.00 & 0.10 & 0.00 \\
        6250  & 0.10 & 0.10 & 0.10 & 0.05 & 0.50 & 0.15 \\
        18750 & 0.05 & 0.10 & 0.05 & 0.10 & 0.35 & 0.35 \\
        40000 & 0.10 & 0.10 & 0.20 & 0.10 & 0.30 & 0.20 \\
        \bottomrule
    \end{tabular}
\end{table}

\textbf{Hybrid-perbatch.} Same probability distribution as \strathyb{}, but instead of sampling the masking strategy for each sequence, samples a fixed strategy per batch. Less computational overhead than the other hybrid strategies.

\textbf{Hybrid-warmstart.} Continues training of an \stratint{} masking model for an additional 50,000 steps. Samples from the following mixing distribution at each step: $[0.10, 0.20, 0.10, 0.15, 0.20, 0.25]$.

\section{Experimental Setup}
\label{ap:experiments}
Each from-scratch run takes $\approx$5 hours on one RTX A6000 (48\,GiB) at 7.2 it/s under fp16; \strathybws{} adds $\approx$2 hours; one-time IgFold prediction cost $\approx$240 GPU-hours on 8 GPUs in parallel.

\subsection{Evaluation}
To evaluate generative recovery of the antigen-binding loop, we measure exact match CDR3 infilling (CDR3) zero-shot on held-out OAS heavy chains. All remaining metrics are evaluated via linear or bilinear probing. For functional binding, paratope AUPRC (Para) and MCC evaluate the identification of antigen-contacting residues on TDC SAbDab\_Liberis (4.5\,\AA{} contact threshold). To assess structural functions, contact map AUROC and long-range precision-at-L (Cont; sequence separation $\geq$ 24) measure tertiary fold recovery on SAbDab crystal structures (8\,\AA{} $C_\alpha$). The structure probe Spearman $\rho$ (Struct) assesses how linearly the embedding space encodes three-dimensional geometry on AB-Bind and SAbDab structures~\cite{sirin2016abbind}. Finally, developability macro-Spearman $\rho$ (Dev) captures the recovery of critical biophysical properties for therapeutic applications across the five TDC TAP metrics~\cite{raybould2019tap}. 

\subsection{Probe Details}
\label{app:probes}
For each task, a task-specific head is trained on last-layer representations from the frozen encoder, which never receives gradient updates. Heads are optimized with AdamW (weight decay $0.01$) at a learning rate of $10^{-3}$ under a linear warmup ($0.1$) with a cosine schedule, gradient norm clipping set at $1.0$, at full precision. For each task we trained multiple independent probes across at least three seeds. At every epoch, the head is scored on the held-out validation split using that task's early stopping metric (Table~\ref{tab:probe_params}). The best validation checkpoint is retained and used for the test set evaluation. We additionally fit a decision threshold on the binary paratope task validation by maximizing Youden's J for paratope MCC; AUPRC and AUROC are threshold-free. The seeds randomize head initialization, dropout, and minibatch order only: the data split and embeddings are identical across seeds. 

Each probing task has its own fixed split. The paratope and developability probes utilize the default splits provided by TDC. TDC SAbDab\_Liberis is partitioned at random into $716/102/205$ antibody sequences (70/10/20), and TAP is partitioned per property and merged by antibody identifier, yielding 241 sequences (70/10/20). The two structural probes are split by PDB entry at 60/20/20 such that heavy and light chains belonging to the same complex are never separated across splits. The contact map probe utilizes 487 SAbDab crystal structures, assigned $292/97/98$ by entry, yielding $367/121/130$ chains. The structure probe merges AB-Bind and SAbDab into a single pool, deduplicated by entry identifier (487 from SAbDab, 31 from AB-Bind). The final structure split is $309/103/104$ by entry and $407/134/139$ by chain. All splits are deterministic and identical across every masking strategy and pretraining seed. CDR3 infilling is performed over $1,000$ sequences from the pretraining hold-out set.

The paratope teacher is a per-token classifier fine-tuned on top of the pretrained AntiBERTa2 encoder using the same TDC SAbDab\_Liberis partition as the paratope probe. It is fit on the training split only, early-stopped on validation AUPRC, and reported on the test split. The encoder and classification head are optimized jointly with AdamW learning rates of $2\times10^{-5}$ and $5\times10^{-4}$ respectively, batch size 16, cosine schedule with $10\%$ warmup, under class-weighted binary cross-entropy. 
As the teacher and probe call the same split function, no test antibody was labeled by the teacher during teacher training; teacher predictions are computed only over the pretraining set, never over evaluation sequences. No structure teacher is trained in this work. Instead, structures are predicted with IgFold, a pretrained antibody structure predictor. To validate the accuracy of the predictions, we compared predicted neighborhoods against real SAbDab crystal structures. Across 316 heavy chains, a median of $93\%$ of each residue's five predicted nearest $C_\alpha$ neighbors also appear among its five nearest neighbors in the crystal structure ($4\%$ for randomly chosen neighbors), with a median Fv $C_\alpha$ RMSD of $0.92$ \AA{}. IgFold is applied only to pretraining sequences to determine mask placement; any overlap between its training data and our structural benchmarks cannot transmit evaluation labels into the encoder.

\textbf{Probe heads.} Paratope (Para): a per-token linear classifier trained with class-weighted binary cross-entropy, scored by AUPRC and MCC. Contact map (Cont): a bilinear form over residue pair embeddings (upper triangle, amino acid positions only) trained with binary CE, scored by AUROC and long-range precision-at-$L$. Structure probe (Struct): a Hewitt-Manning linear structural probe~\cite{hewitt-manning-2019-structural} that predicts squared $C_\alpha$ distances, trained with masked MSE and scored by the Spearman $\rho$ between predicted and true distances. Developability (Dev): mean pooling over residue tokens followed by a linear regressor trained with MSE on $z$-scored targets, scored by the macro-averaged Spearman $\rho$ over five TAP properties (encoder hidden size is $512$ dimensions).

 \begin{table}[h]
    \caption{Linear-probe configurations. Patience $0$ disables early stopping
  (developability trains its full budget).}
    \label{tab:probe_params}
    \centering
    \footnotesize
    \resizebox{0.8\textwidth}{!}{%
    \begin{tabular}{lccccl}
      \toprule
      Task & Epochs & Batch & Patience & Seeds & Early-stop metric \\
      \midrule
      Paratope        & 50  & 32 & 10 & 3 & AUPRC \\
      Contact map     & 50  & 16 & 10 & 3 & precision@$L$ \\
      Structure probe & 100 & 4  & 15 & 3 & Spearman $\rho$ \\
      Developability  & 100 & 16 & 0  & 5 & macro Spearman $\rho$ \\
      \bottomrule
    \end{tabular}}
  \end{table}

\end{document}